\documentclass{evolang12}
\usepackage{graphicx}
\usepackage{url}
\usepackage{authblk}

\begin{document}

\title{Compositional properties of emergent languages in deep learning}
\author[1]{Bence Keresztury}
\author[2]{Elia Bruni}
\affil[1]{University of Amsterdam, kbence16@gmail.com}
\affil[2]{Universitat Pompeu Fabra,  elia.bruni@upf.edu}



\maketitle

\abstracts{Recent findings in multi-agent deep learning systems point towards the emergence of compositional languages. These claims are often made without exact analysis or testing of the language. In this work, we analyze the emergent language resulting from two different cooperative multi-agent game with more exact measures for compositionality. Our findings suggest that solutions found by deep learning models are often lacking the ability to reason on an abstract level therefore failing to generalize the learned knowledge to out of the training distribution examples. Strategies for testing compositional capacities and emergence of human-level concepts are discussed.}

\section{Introduction}

In the field of deep learning the study of emergent languages became a hot-topic in the last couple years. Usually two or more agents play a cooperative game where agents' goals are common but they access different information. In order to solve the given task agents have to share useful information with each other through a discrete bottleneck called the communication channel. The discrete symbols in the message do not have any a priori meaning but agents learn to cooperate by attributing meaning to the messages; a language protocol emerges as a by-product of the training process. This emergent language serves only one goal: to complete the task successfully. One of the promises of this approach is to provide meaningful insights into the early stages of human language emergence as a result of cooperation.

One key aspect of human languages is the compositional structure which allows to describe potentially infinite amount phenomena with a finite vocabulary. Compositional learning is the ability to abstract and store rules and skills that are reusable for different tasks. On the contrary, today's neural networks are extremely sample inefficient, requiring huge amounts of training data. In the deep learning literature some researchers claim agents to have developed compositional languages. However, this compositionality is often claimed by either eye-balling the language or trying to hand-label symbols based on the usages in a confined setting. 

One good measure of compositional thinking is the ability to systematically generalize to new data distributions. If an agent's understanding of the world is fully compositional they must be able to recombine the knowledge units they've learned to make appropriate decisions in scenarios where the units of knowledge are recombined in a previously unseen fashion. A compositional language therefore would allow agents to talk about combinations of objects that they have never seen together before. In this work we want to test these claims and the level of compositionality that these languages possess.

\section{Related Work}

The study of emergent languages in multi-agent deep learning setup begin with initial work of \citeA{lazaridou2016multi} and since then it became a widely-studied phenomenon in the field. The original setting contains two agent, a Sender who sees an image and a Receiver who sees the same image alongside with a distractor. The Sender submits a character out of a well-defined set of vocabulary, while the receiver makes a decision about which picture could have been seen by the Sender based on its message. The agents get a shared reward at the end of every trial therefore it is a purely common interest game. The authors show that agents are able to successfully cooperate and develop a meaningful communication protocol. However, \citeA{bouchacourt2018agents} argued that the feature representations induced by the task (Sender and Receiver representations) do not capture the conceptual properties of the depicted objects, agents are more likely to concentrate on low-level relational properties based on what agents are able to successfully distinguish between images but no semantically meaningful communication occurs. That means their understanding does not align with human intuition about what is on the picture, therefore their communication will not be about abstract, human-level concepts.

Subsequent works concentrated on making the task more challenging and the communication process more human-like by using raw pixels as input and a sequence of symbols as the message \cite{lazaridou2018emergence}, allowing agents to take multiple communication steps before making a decision \cite{evtimova2017emergent} or using a community of agents to make the emergent language less idiosyncratic \cite{tieleman2018shaping}.   

Since one of the distinguishing properties of human languages is their compositional structure, researchers often examine the emergent language in order to find compositional patterns. \citeA{Havrylov2017} shows examples where images that generate messages with the same prefix belong to the same semantic category. They argue therefore that agents use a hierarchical coding of characteristics to describe images. \citeA{Mordatch2017EmergenceOG} used emergent communication between agents in a simulated two-dimensional environment where different agents have different goals (going to landmarks, convince other agents to go to landmarks or to other agents, etc.) but their rewards are shared. They claim compositionality of the resulting language by manually labelling symbols of their emergent language and showing emergence of the concepts of \textit{goto} or agent identities. Even though these concepts seem to be consistent through different trials, hand-labelling is only feasible for languages with a very limited set of vocabulary. 

Independently from emergent language literature, a line of research aims to test the generalization abilities of deep learning models. While being able to generalize to previously unseen scenarios is one of the core aspects of human intelligence \cite{minsky1988society}, a lot of influential thinkers argued that connectionist models of cognition cannot account for the observed systematic compositionality due to their associative nature \cite{fodor1988connectionism}. Recent findings corroborate these claims by showing that modern generic purpose Neural Network models struggle to extract syntactic rules from training data, thus they generalize poorly to previously unseen tasks and data distributions (e.g. \citeA{lake2017generalization} or \citeA{loula2018rearranging}). This happens because neural networks often latch on to dataset-specific regularities instead of distilling syntactic rules in form of logical formulas \cite{bahdanau2018systematic}.

\section{Method}

\subsection{Visual Question Answering}

To test the generalization properties of different models \citeA{bahdanau2018systematic} created the Spatial Queries on Object Pairs (SQOOP) data set, a visual question answering task, where the main challenge is the necessity to generalize to previously unseen combination of known objects and relations. This task is easy for a human to do since we can easily decide whether the statement "\textit{there is a golf-ball under the hat}" is true or not even if we have never seen the combination of \textit{golf-ball} and \textit{hat} in one sentence before. 

\begin{figure}[ht]
\begin{center}
\scalebox{0.6}{\includegraphics{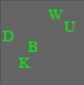}}
\end{center}
\caption{{\footnotesize An example image of our SQOOP dataset \label{sqoop_example}}}
\end{figure}

The pictures are 64x64x3 pixel RGB images with five characters on each. A question is asked about the arrangement of the letters which is either true or false (for an example see Figure \ref{sqoop_example}). The questions in the training set are constrained to only a subset of the possible character combinations, so the test set can assess how well an agent is able to disentangle important features of the input to understand objects and relational concepts. That means for example that the training set only contains questions about \textit{A} in relation to \textit{B} (e.g. \textit{Is A left of B?}) but the test set contains questions about \textit{A} in relation to other characters (e.g. \textit{Is A above C?}). The task becomes increasingly harder as the training set gets more constrained to fewer character combinations. 

\subsection{Referential games}
Referential games are a specific type of cooperative multi-agent games that is a popular choice among researchers of the emergent language phenomena. The game is structured in the following way:

\begin{enumerate}
    \item There are $K$ agents, each parameterized by $\Theta_k$
    \item There is a set of images tuples $\{(o_1, t_1, d_{1_1}, ..., d_{1D}), (o_2, t_2, d_{21}, ..., d_{2D}) ..., \\ (o_N, t_N, d_{N1}, ..., d_{ND})\}$ where $o$ denotes the original images that the Sender agents see, $t$ denotes the target image and $d$ denotes the distractor images.
    \item There is a vocabulary $V$ consisting of $|V|$ discrete units ("words") and a message $M$ is created from $L$ of these words, where $L$ is the length of the message.
    \item In every round, there are two agents playing, a Sender and a Receiver. The Sender $s$ gets an original image as an input and omits a message based on his policy.
    $m = s_{\Theta_s}(o) \in (V\times V\times...\times V)$
    \item The Receiver gets the message $m$, the distractor images $d_1, d_2, ..., d_D$ and the target image $t$ as an input. However, it does not see the original image $o$ that the Sender agent sees. The target image is mixed between the distractor images and the Sender task is to point to the target by outputting it's position in the order of images, that means 
    $r_{\Theta_r}(m, d_1, ..., d_j, t, d_{j+1}, ..., d_D) = j+1$ for the correct answer.
    
\end{enumerate}{}

In the most simple case there are only two agents in the population ($K=2$), 1 distractor image ($D=1$) and the original image is the same as the target image ($o=t$). In our task there has been 2 agents, 2 distractors for each target and, in order to encourage agent's to communicate about higher level properties, the target image is perturbed ($o \neq t$). 

We adapted the Visual Question Answering SQOOP dataset to the referential game. We created the target image by preserving all spatial relations between the characters while on distractor images at least one of the original relationships does not hold. Examples can be seen on Figure. Notice that visually the distractors are very close, making it challenging for deep learning models to rely on low-level features.
\begin{figure}[ht]
\begin{center}
\scalebox{0.4}{\includegraphics{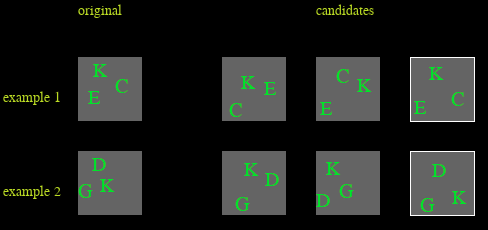}}
\end{center}
\caption{{\footnotesize Two samples from the SQOOP referential game. The target is highlighted \label{ex_ref}}}
\end{figure}

\subsection{Agents' architecture}

The Sender's architecture was identical in both games. The image has been processed through a Convolutional Neural Network \cite{lecun1995convolutional}, which resulted in the Sender's representation after a linear layer. A Long Short-term Memory Network or LSTM for short \cite{hochreiter1997long} processed the message tokens one by one, using a standard embedding layer for every word to initialize the next word's production. The message has been processed through the Receiver's LSTM network and passed through a linear layer. In the Visual Question Answering game, the question has been processed by a different LSTM and the two streams of information has been combined by a FiLM layer \cite{perez2018film}. The final binary decision has been made by a multi-layer fully-connected Neural Network. In the referential game the candidate pictures has been fed through the same Convolutional Neural Net and the decision has been made by taking the maximum of the dot product of these activation with the activations coming from the communication stream. For exact hyperparameter settings and training details we refer to the supplementary material.

\section{Experiments}

\subsection{Visual Question-Answering task}

\begin{figure}[ht]
\begin{center}
\scalebox{0.25}{\includegraphics{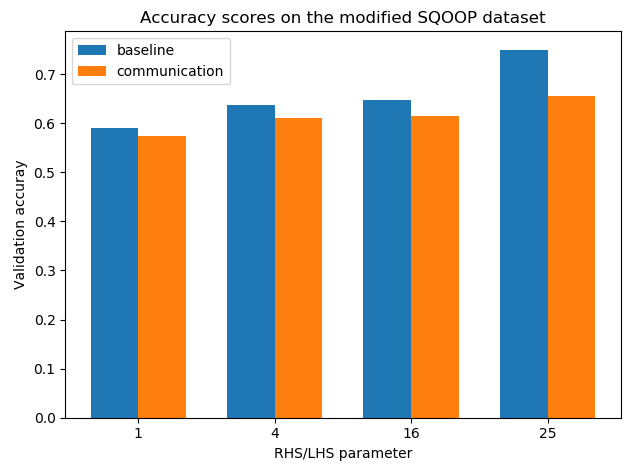}}
\end{center}
\caption{{\footnotesize Comparison of generalization capabilities \label{vqa}}}
\end{figure}

Firstly, we used the visual question-answering task to see the systematic generalization properties of agents. We compared the baseline, single agent's generalization capabilities to the multi-agent setting. The latter shows how easily agents can use the emergent language to talk about new unseen data distributions. The results can be seen on the Figure \ref{vqa}. 

The RHS/LHS parameter shows how many characters agents see in the training set with every character on the left side, that means the task becomes harder when the value is smaller. We see that the baseline model is able to generalize better to unseen distributions. On the other hand we inspect that the performance of both models is relatively low. This can come from the fact that we unbiased the data in a sense that neither the images nor the questions contain any information about the label of the example which makes the task even harder (for the exact process see supplementary materials). The other difficulty of the task was that in order to be able to answer any question, agents had to convey information about all possible pairs of 5 characters that were on the image. With less characters on the image, we saw the performance increase radically. 

\subsection{Referential game and language analysis}

\begin{figure}[ht]
\begin{center}
\scalebox{0.3}{\includegraphics{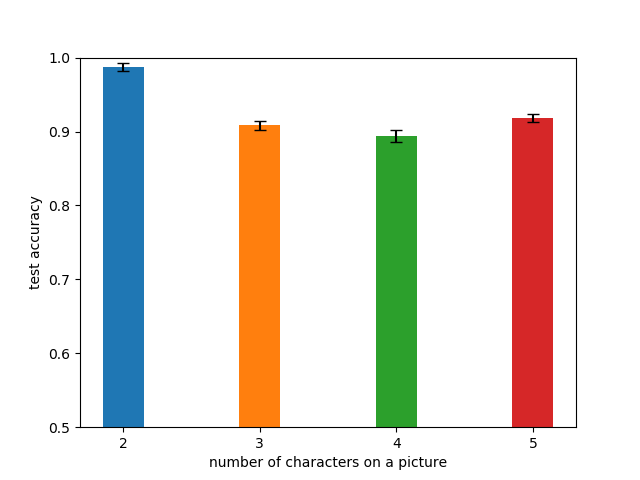}}
\end{center}
\caption{{\footnotesize Referential game accuracy scores. Average and standard deviation of 3 runs are plotted. \label{ref_acc}}}
\end{figure}

The accuracy scores on the referential game framework are depicted on Figure \ref{ref_acc}. We see that agent's are able to play the game around 90\% accuracy on all of the conditions. 

To test properties of the emerged languages are we conducted further analysis on them. We show couple of examples messages on Table \ref{tab:example_messages}.

\begin{table}
\tablecaption{Example messages}
{\footnotesize
\begin{tabular}{llllllllllllllll}
\hline
m1: & 32 & 27 & 32 & 6  & 32 & 6  & 32 & 27 & 32 & 45 & 32 & 32 & 45 & 26 & 26 \\ \hline
m2: & 9  & 9  & 27 & 21 & 27 & 21 & 6  & 27 & 32 & 9  & 6  & 21 & 6  & 26 & 32 \\ \hline
m3: & 27 & 27 & 21 & 6  & 6  & 9  & 6  & 27 & 27 & 6  & 9  & 6  & 9  & 9  & 27 \\ \hline
m4: & 6  & 6  & 26 & 21 & 32 & 45 & 32 & 9  & 32 & 32 & 27 & 32 & 6  & 6  & 32 \\ \hline
\end{tabular}}
\label{tab:example_messages}
\end{table}

Agents communicate with integer numbers but to stress the analogy with human languages we will refer to the number as words. We can see that certain words dominate the messages. Furthermore we see a couple of other interesting phenomena. Agents tend to repeat words (e.g. in message 1 \textit{32} \textit{32}) or use 2-3 words often together. For example the pattern \textit{27} \textit{32} is there in message 1, 2 and 3, in message 1 and 3 it is preceded by \textit{32}). Such collocations are also characteristic of human languages (e.g. doing homework, saving time, ...). In order to avoid cherry-picking of examples, we provide distributional properties in the supplementary material. 

Topographical similarity shows how similar two representational spaces are. That means we can examine whether vectors that are close to each other in one space are staying close in an other or not. It is calculated as the correlation of the distances between all possible pairs of representations in two spaces. We measured distances between representations as the cosine distance between activation vectors.

\begin{figure}[ht]
\begin{center}
\scalebox{0.3}{\includegraphics{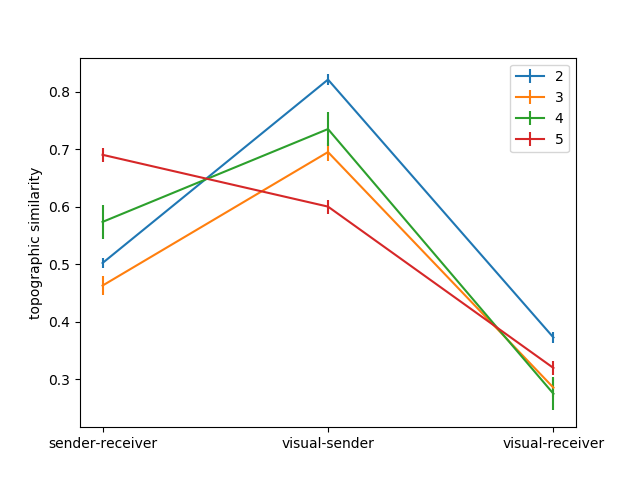}}
\end{center}
\caption{{\footnotesize Representational similarity analysis \label{rsa}}}
\end{figure}

On Figure \ref{rsa} we see that similarity scores are all positive which is what we expected. The sender-receiver topographical similarity measures how different the two agents represent the same input. A completely successful communication protocol would mean that the two representations are completely aligned (e.g. all information is conveyed), the information is able to be compressed and passed through the discrete bottleneck. We see however, that this is not the case; even if the topology of the two spaces are similar in all conditions they are far from being perfectly aligned. The most successful representation alignment can be observed in the task with the most characters on the picture. This corroborates our suspicion that agents might communicate about lower-level features, since in theory, agents have to compress the most amount of information in the case of 5 characters per image, where they seem to be the most successful. The representational structure between the visual representation and the Sender agent is fairly well preserved, while the topological space becomes increasingly dissimilar between the visual and the Receiver's representation. This is also understandable since a lot of information is getting lost in the discrete bottleneck.

Finally, we trained diagnostic classifiers to decide whether all character-level information is encoded in the messages. We do this as a check for compositionality. We humans would solve the task by trying to communicate about the position of the characters, therefore we could easily export our knowledge to unseen character combinations. If agents communicate about the presence of characters as well, we must be able to recover the identity of the characters by training a classifier with the messages as inputs. We see however, that this is not the case. We weren't able to recover the identity of the characters which means agents must communicate about lower-level properties and do not come up with the disentangled representation of characters. Moreover, if we check different representational spaces in the model's architecture we see, in accordance with the topological similarity score, that the information is gradually lost in the system. Even the visual features are not fully capturing conceptual properties of the images therefore the model is doomed to talk on a less abstract, less compositional level, less human-like level. 

\begin{figure}[ht]
\begin{center}
\scalebox{0.3}{\includegraphics{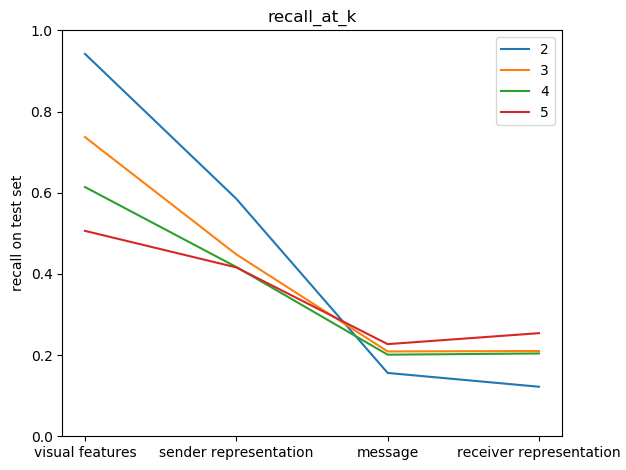}}
\end{center}
\caption{{\footnotesize Recall at k. How many characters of the top k activations are on the original image. Lines for different number of characters per image are plotted. \label{reck}}}
\end{figure}

\section{Discussion}

In our first experiment we saw that induced bottleneck in the system did not help to systematically reuse knowledge in neither of the conditions. This also means that multi-agent communication has proven to be rather a hindrance for learning than a useful tool for better generalization. This failure shows that an emergence language in itself is not sufficient to enforce compositional learning and the emergence of systematic generalization properties. We think that some form of extra supervision on the emergent language is necessary to encourage truly compositional thinking and reusable knowledge representation. Furthermore, we saw how task success on a task that humans solve compositionally does not mean the emergence of abstract human-level concepts. From the messages we could not reconstruct the presence of all characters in neither of the conditions. Even their visual representation did not evolve to capture all high-level information about characters. 

We believe that research of emergent languages should contain more thorough metrics of compositionality and compulsory testing on systematic generalization task before claiming compositional properties of the emergent language. More thorough analysis of learned representations often point to the fact that deep learning models latch onto and exploit dataset-specific regularities instead of learning systematic solutions to problems. We believe that testing on out of distribution generalization tasks and diagnostic classifiers are a good way to uncover these biases in the system and creating a more insightful research direction into the emergence of compositional languages thus contributing more to the understanding of the early stage human language evolution. 

\section{Acknowledgements}
EB is funded by the European Union’s Horizon 2020 research and innovation program under the Marie Sklodowska-Curie
grant agreement No 790369 (MAGIC).

\bibliographystyle{apacite}
\bibliography{evolang12} 

\end{document}